\documentclass[conference]{IEEEtran}
\ifCLASSINFOpdf
   \usepackage[pdftex]{graphicx}
\else
\fi
\usepackage{amsmath}
\usepackage[utf8]{inputenc} %TODO MW dzialaja polskie znaki/tekst
\usepackage[T1]{fontenc}
\usepackage{subcaption}

\begin{document}
%TODO Do przemyślania
\title{Vision based hardware-software real-time control system for autonomous landing of an UAV}

\author{
	\IEEEauthorblockN{Krzysztof Blachut}
	\IEEEauthorblockA{ AGH University of Science \\
		and Technology Krakow, Poland \\
		E-mail: kblachut@agh.edu.pl}
	\and
	\IEEEauthorblockN{Hubert Szolc}
	\IEEEauthorblockA{ AGH University of Science \\
		and Technology Krakow, Poland \\
		E-mail: szolc@agh.edu.pl}
	\and
	\IEEEauthorblockN{Mateusz Wasala}
	\IEEEauthorblockA{ AGH University of Science \\
		and Technology Krakow, Poland \\
		E-mail: wasala@agh.edu.pl}
	\and
	\IEEEauthorblockN{Tomasz Kryjak, \textit{Senior Member IEEE}}
	\IEEEauthorblockA{ AGH University of Science \\
		and Technology Krakow, Poland \\
		E-mail: tomasz.kryjak@agh.edu.pl}
	\and
	\IEEEauthorblockN{Marek Gorgon, \textit{Senior Member IEEE}}
	\IEEEauthorblockA{ AGH University of Science \\
		and Technology Krakow, Poland \\
		E-mail: mago@agh.edu.pl}
}


%Telephone: (800) 555--1212, Fax: (888) 555--1212}


\maketitle

\begin{abstract}

	In this paper we present a~vision based hardware-software control system enabling autonomous landing of a~multirotor unmanned aerial vehicle (UAV).
	It allows the detection of a~marked landing pad in real-time for a~1280~x~720 @ 60 fps video stream.
	In addition, a~LiDAR sensor is used to measure the altitude above ground.
	A~heterogeneous Zynq SoC device is used as the computing platform.
	The solution was tested on a~number of sequences and the landing pad was detected with 96\% accuracy. 
	This research shows that a~reprogrammable heterogeneous computing system is a~good solution for UAVs because it enables real-time data stream processing with relatively low energy consumption.

	%W artykule przedstawiono sprzętowo-programowy system umożliwiający autonomiczne lądowanie bezzałogowego pojazdu latającego.
	%Jego zasadniczym elementem jest działający w czasie rzeczywistym, dla strumienia o~rozdzielczości 1280x720 @ 60 fps, system wizyjny, który pozwala na detekcję oznaczonego lądowiska.
	%Dodatkowo, do pomiaru wysokości wykorzystano czujnik LiDAR.
	%Jako platformę obliczeniową zastosowano heterogeniczny układ Zynq SoC.
	%Rozwiązanie zostało przetestowane na szeregu sekwencji -- uzyskano skuteczność XX.
	%Przeprowadzone eksperymenty pokazują, że reprogramowalne heterogeniczne układy obliczeniowe stanowią dobre rozwiązanie dla UAV, gdyż umożliwiają przetwarzanie strumienia danych w czasie rzeczywistym przy relatywnie niewielkim zużyciu energii.
	
	%In this paper we present a vision based hardware-software control system enabling autonomous landing of a multirotor unmanned aerial vehicle (UAV). It allows the detection of a marked landing pad in real-time for a 1280 x 720 @ 60 fps video stream. In addition, a LiDAR sensor is used to measure the altitude above ground. A heterogeneous Zynq SoC device is used as the computing platform. The solution was tested on a number of sequences and the landing pad was detected with 96% accuracy. This research shows that a reprogrammable heterogeneous computing system is a good solution for UAVs because it enables real-time data stream processing with relatively low energy consumption.	

\end{abstract}


% \fi
\IEEEpeerreviewmaketitle

% --------------------------------------------------------------------------------------------------------
\section{Introduction}

Unmanned Aerial Vehicles (UAV), commonly known as drones, are becoming increasingly popular in commercial and civilian applications, where they often perform simple, repetitive tasks such as terrain patrolling, inspection or goods delivering.
Especially popular among drones are the so-called multirotors (quadrocopters, hexacopters etc.), which have very good navigation capabilities (vertical take-off and landing, hovering, flight in narrow spaces).
Unfortunately, their main disadvantage is high energy consumption. 
With the currently used Li-Po batteries the flight time is relatively short (up to several dozen minutes). 
Therefore, autonomous realization of the mentioned tasks requires many take-offs and landings to replace or recharge batteries.

%Bezzałogowe pojazdy latające, zwane potocznie dronami, zyskują coraz większą popularność w zastosowaniach komercyjnych i cywilnych, gdzie bardzo często wykonują proste, powtarzalne zadania np. patrolowanie terenu, czy dostarczanie przesyłek.
%Wśród dronów szczególnie popularne są tzw. wielowirnikowce (quadrocoptery, hexacoptery), które charakteryzują się bardzo dobrymi możliwościami nawigacyjnymi. 
%Niestety ich podstawową wadą jest duże zużycie energii, co przy obecnie stosowanych akumulatorach przekłada się na stosunkowo krótki czas lotu (kilkanaście minut). 
%Autonomiczna realizacja wspominanych zadań wymaga zatem przeprowadzenia wielu startów i~lądowań.

The start of a~multirotor, assuming favourable weather conditions (mainly lack of strong wind) and while the distance from other obstacles is preserved, is simple.
Landing in a~selected place, however, requires relatively precise navigation.
If a~tolerance of up to several meters is allowed, the~GPS (Global Positioning System) signal can be used.
Nevertheless, it should be noted that in the presence of high buildings the GPS signal may disappear or be disturbed.
What is more, even under favourable conditions, the accuracy of the determined position is limited to approximately 1--2~m.
Performing a~more precise landing requires an additional system.
There are primarily two solutions to this problem -- the first is based on computer vision and the second uses a~radio signal to guide the vehicle.

%Start wielowirnikowca, przy założeniu sprzyjających warunków atmosferycznych (głównie braku mocnego wiatru) oraz pewnego oddalenia od potencjalnych przeszkód, jest prosty.
%Natomiast wylądowanie w~określonym miejscu wymaga względnie precyzyjnej nawigacji.
%Jeśli dopuszcza się tolerancję do kilku metrów, to można wykorzystać sygnał GPS.
%Należy jednak zaznaczyć, że w warunkach obecności wysokiej zabudowy może on zanikać.
%Ponadto nawet w sprzyjających warunkach dokładność wyznaczonej pozycji jest ograniczona do 1-2 m.
%Wykonanie bardziej precyzyjnego lądowania wymaga dodatkowego systemu.
%Spotyka się dwa rozwiązania -- oparte o system wizyjny oraz naprowadzanie sygnałem radiowym.

The main contribution of this paper is the proposal of a~hardware-software vision system that enables precise landing of the drone on a~marked landing pad.
As the computational platform, a~heterogeneous Zynq SoC (System on Chip) device from Xilinx is used.
It consists of programmable logic (PL or FPGA -- Field Programmable Gate Array) and a~processor system (PS) based on a~dual-core ARM processor.
This platform allows to implement video stream processing with~$ 1280 \times 720 $ resolution in real time (i.e. 60 frames per second) in programmable logic, with relatively low power consumption (several watts).
The processor facilitates communication with the drone controller and is responsible for the final part of the vision algorithm.
To the best of our knowledge, this is the first reported implementation of such a~system in a~heterogeneous device.

%Najważniejszym osiągnięciem pracy jest propozycja sprzętowo-programowego systemu wizyjnego, który umożliwia precyzyjne wylądowanie drona na oznaczonym lądowisku.
%Jako platformę obliczeniową zastosowano heterogeniczny układ obliczeniowy Zynq SoC (System on Chip) firmy Xilinx.
%Składa się on z~logiki programowalnej (układ FPGA (Field Programmable Gate Array)) oraz systemu procesorowego opartego o procesor ARM.
%Użycie takiej platformy pozwala na realizację przetwarzania sygnału wideo o~rozdzielczości $1280 \times 720$ w czasie rzeczywistym (tj. 60 fps) w zasobach rekonfigurowalnych, przy stosunkowo niewielkim zużyciu energii.
%Natomiast procesor znacząco ułatwia komunikację ze sterownikiem drona oraz odpowiada za końcową część algorytmu wizyjnego.
%Zgodnie z wiedzą autorów jest to pierwsze tego typu rozwiązanie opisane w~literaturze.

The reminder of this paper is organized as follows. 
In Section~\ref{sec:prev_work} previous works on autonomous drone landing are briefly presented. 
Then, the multirotor used in the experiments, landing procedure and landing pad detection algorithm are described. 
In Section~\ref{sec:hw_sw} the details of the created hardware-software system are presented. 
Subsequently, in Section~\ref{sec:evaluation}, the evaluation of the designed system is discussed. 
The paper concludes with a~summary and further research directions discussion.

%The reminder of this paper is organized as follows.
%W rozdziale \ref{sec:prev_work} omówiono dotychczasowe prace nt. autonomicznego lądowania drona.
%Następnie, w~rozdziale \ref{sec:algorithm} opisano wykorzystanego w eksperymentach drona, procedurę lądowania oraz algorytm detekcji i śledzenia lądowiska.
%W~rozdziale \ref{sec:hw_sw} przedstawiono szczegóły stworzonego systemu sprzętowo-programowego.
%W~rozdziale \ref{sec:evaluation} zaprezentowano wyniki ewaluacji systemu.
%Artykuł kończy podsumowanie i~wskazanie dalszych kierunków badań. 

% --------------------------------------------------------------------------------------------------------
\section{Previous works}
\label{sec:prev_work}

The topic of the autonomous landing with the use of visual feedback has been addressed in a~number of scientific papers.
In the following review we focus mainly on the landing on a~static pad.

% Temat autonomicznego lądowania z~wykorzystaniem wizyjnego sprzężenia zwrotnego został poruszony w szeregu prac naukowych.
% W niniejszym przeglądzie skupiono się głównie na pracach dotyczących lądowania na lądowisku statycznym.

The authors of the work \cite{Jin_2016} reviewed different autonomous landing algorithms of a~multirotor on both static and moving landing pads.
They compared all previously used landing pad tags, i.e.~ArUco, ARTag, ApriTag, tags based on circles and a~``traditional'' H-shaped marker.
They also pointed out that landing in an unknown area, i.e.~without a~designated landing pad, is a~challenging task.
Then they analysed two types of drone controllers during the landing phase: PBVS (Position-Based Visual Servoing) and IBVS (Image-Based Visual Servoing).
The first compares drone's position and orientation with the expected values.
The second one compares the location of feature points between the pattern and subsequent image frames.
In the summary, the authors pointed out three main challenges related to this type of systems: development of a~reliable vision algorithm with limited computing resources, improvement of a~state estimation and appropriate modelling of the wind in the control algorithm.

% Autorzy pracy \cite{Jin_2016} przeprowadzili przegląd dotychczas zaproponowanych metod związanych z autonomicznym lądowaniem wielowirnikowca zarówno na statycznym lądowisku, jak i na platformie mobilnej. 
% Porównano wszystkie wykorzystywane do tej pory oznaczenia lądowiska tj. ArUco, ARTag, ApriTag oraz bazujące na
% okręgach lub znaczniku w~postaci litery H. 
% Wskazano również, że wyzwanie stanowi lądowanie w~nieznanym terenie tj. bez oznaczonego lądowiska.
% Następnie analizie poddano kontrolery sterowania dronem podczas lądowania. 
% Wyróżniono dwa rodzaje regulatorów: PBVS (Position-Based Visual Servoing) i
% IBVS (Image-Based Visual Servoing). 
% Pierwszy z nich porównuje wyznaczoną pozycję i orientację drona z wartościami oczekiwanymi. 
% Natomiast drugi porównuje położenie punktów charakterystycznych pomiędzy wzorcem, a~kolejnymi ramkami obrazu.
% Autorzy w~podsumowaniu wskazują trzy wyzwania związane z~tego typu systemami: stworzenie niezawodnego algorytmu wizyjnego, przy ograniczonych zasobach obliczeniowych, poprawę estymacji stanu oraz odpowiednie uwzględnienie wiatru w~algorytmie sterowania.

In the work \cite{Qiu_2017} the authors proposed an~algorithm for landing of an~unmanned aerial vehicle on a~stationary landing pad.
It was used when replace or re-charge the drone's battery was necessary during high-voltage line inspection.
For the detection of the landing pad they used thresholding, median filtration, contour extraction, determination of geometrical moments and an~SVM classifier.
In addition, the Extended Kalman Filter (EKF) was used to determine the position of the drone. 
It processed data from inertial sensors (IMU -- Inertial Measurement Unit), radar and vision system.
During tests, the drone approached the landing pad with a~GPS sensor and then switched to the vision mode, in which the landing pad was detected with a~camera.
Out of 20 sequences, 14 ended with a~position error of less than 10~cm, and the remaining ones below 20~cm.
At the same time, in 12 tests the orientation error was below 10 degrees, while in the remaining ones below 20 degrees.
As a~computing platform, the authors used Raspberry Pi.
They obtained a~processing time of a~single frame of 0.19~s, that is about 5~frames per second, which is sufficient for slow-moving drones (no information about the camera resolution was provided).

% W pracy \cite{Qiu_2017} autorzy zaproponowali algorytm lądowania bezzałagowego pojazdu latającego na nieruchomym lądowisku. 
% Wywołanie tej procedury ma na celu wymianę lub podładowania źródła energii znajdującego się na dronie w~trakcie inspekcji linii wysokiego napięcia. 
% Do detekcji lądowiska wykorzystano kolejno binaryzację, filtrację medianową, ekstrakcję konturu, wyznaczenie momentów geometrycznych oraz klasyfikator SVM. 
% Do określenia pozycji drona zastosowano rozszerzony filtr Kalmana (EKF), który przetwarzał dane pochodzącego z czujników inercyjnych (IMU), radaru oraz systemu wizyjnego. 
% Testy opierały się na zbliżeniu się drona do lądowiska za pomocą czujnika GPS, a następnie przełączeniu na tryb wizyjny i detekcji lądowiska za pomocą kamery. 
% Na 20 przeprowadzono prób 14 zakończyło się błędem pozycji poniżej 10cm, pozostałe poniżej 20cm, zaś 12 prób błędem orientacji poniżej 10 stopni, pozostałe poniżej 20 stopni.
% Jako platformę obliczeniową autorzy użyli Raspberry Pi.
% Uzyskali czas przetwarzania pojedynczej ramki 0.19 s tj. ok. 5 ramek na sekundę (brak informacji o~rozdzielczości kamery), co jest wystarczające dla wolno poruszających się dronów.

The authors of the work \cite{Xu_2018} presented an~algorithm for controlling autonomous landing of an~unmanned aerial vehicle on a~stationary T-shaped landing pad.
The proposed vision algorithm was based on so-called image key points (feature points).
The algorithm consisted of colour to greyscale transformation, thresholding, morphological operations, contour extraction (by using the Laplace operator) and matching the polygon to the detected points.
Based on this, 8~angular points were determined, which were used to find the position of the drone corresponding to the landing pad.
The algorithm worked in real-time as the processing of one frame with a~resolution of $ 480 \times 320 $ pixels took about 50~ms.
The authors did not state on what platform the algorithm was implemented.

% Autorzy pracy \cite{Xu_2018} przedstawili algorytm autonomicznego lądowania bezzałogowego pojazdu latającego na nieruchomym lądowisku w kształcie litery T.
% Zaproponowany wizyjny algorytm nawigacyjny oparty jest na znalezieniu punktów charakterystycznych na obrazie.
% Składa się on z transformacji obrazu do odcieni szarości, binaryzacji, operacji morfologicznych, ekstrakcji konturu (the Laplace operator) i dopasowania wielokąta do wykrytych punktów. 
% Na podstawie tego wyznaczanych jest 8 wierzchołków (angular points).
% Te punkty służą do określenia pozycji drona względem lądowiska.
% Algorytm działa w czasie rzeczywistym.
% Czas przetwarzania jednej ramki o rozdzielczości 480 x 320 pikseli zajmuje około 50ms.
% Brak informacji o użytej platformie obliczeniowej.

The article \cite{Lee_2018} presents the possibility of using reinforcement learning to generate the control necessary for autonomous landing.
In the learning phase a~simulation environment was used.
Real experiments were also carried out on a~hexacopter.
It was equipped with a~custom computer based on a~TI microcontroller, NVIDIA Jetson TX2 platform and HD camera.
The authors reported a~single frame processing time of 200 ms (5~fps).
In addition, they mentioned that this was one of the reasons for the observed oscillations. %TODO ew. w kolejnej edycji poprawić.

% W~artykule \cite{Lee_2018} zaprezentowano możliwość zastosowania reinforcement learing do generowania sterowania niezbędnego przy autonomicznym lądowaniu. 
% W fazie uczenia zastosowano środowisko symulacyjne.
% Przeprowadzono również eksperymenty na rzeczywistym heksacopterze.
% Był on wyposażony w custom komputer oparty o mikrokontroler TI, platformę NVIDIA Jetson TX2 oraz kamerę o rozdzielczości HD.
% Autorzy wspominają o czasie przetwarzania pojedynczej ramki wynoszącym 200 ms (5 fps).
% Dodatkowo wspominają, że jest to jedna z przyczyn zaobserwowanych oscylacji.

The article~\cite{Patruno_2019} presents a~complex vision algorithm for landing pad detection.
A~marker in the form of the letter~H placed inside a~circle with an additional smaller circle in the centre was used.
The algorithm operation was divided into three phases that were performed depending on the distance between the drone and the landing pad.
In the first one, the outer circle, then the letter~H, and finally the middle circle were detected.
The authors devoted much attention to the reliability of the algorithm, providing correct operation in the conditions of partial occlusion and shading of the marker.
Odroid XU4 computing platform with a~Linux operating system was used and the OpenCV library was applied.
The source of the video stream was a~camera with a~resolution of $ 752 \times 480 $ pixels.
Processing of 12 to 30 frames per second was obtained depending on the phase of the considered algorithm.

% W~artykule \cite{Patruno_2019} zaprezentowano złożony algorytm wizyjny do detekcji lądowiska.
% Zastosowano znacznik w~postaci listery H umieszczonej wewnątrz okręgu z~dodatkowym mniejszym okręgiem w~środku.
% Działanie algorytmu podzielone jest na trzy fazy w zależności od odległości pomiędzy dronem, a lądowiskiem.
% W pierwszej wykrywane jest zewnętrzny okrąg, następnie litera H, a w ostatniej środkowy okrąg (wtedy pozostałe mogą nie być widoczne).
% Autorzy dużą uwagę poświęcili niezawodności algorytmu,  w tym poprawnemu działaniu w warunkach częściowego przesłonięcia i zacienienia znacznika.
% Jako platformę obliczeniową zastosowano OdroidXU4, z uruchomionym systemem operacyjnym typu Linux.
% Wykorzystano bibliotekę OpenCV.
% Źródłem obrazu była kamera o~rozdzielczości 752 x 480 pikseli.
% Uzyskano przetwarzanie od 12 do 30 ramek na sekundę w~zależności od rozpatrywanej fazy algorytmu.  

In~the work \cite{Huang_2019} the authors proposed the RTV (Relative Total Variation) method to filter the unstructured texture to improve marker (a~triangle inside a~circle) detection reliability.
They also used median filtering, Canny edge detection and polygon fitting by the Ramer-Douglas-Peucker algorithm.
In addition, they proposed a~method of integrating data obtained from the GPS sensor and vision algorithm.
There was no information about used computing platform, camera resolution or the possibility of performing calculations in real-time. 

% W~pracy \cite{Huang_2019} zaproponowano użycie metody RTV (Relative Total Variation) do odfiltrowania nieustrukturyzowanej tekstury, co pozwala poprawić niezawodność detekcji markera.
% Ponadto użyto filtracji medianowej, detekcji krawędzi metodą Canny oraz dopasowania wielokąta algorytmem Ramer-Douglas-Peucker (znacznik ma postać trójkąta wewnątrz okręgu).
% Dodatkowo zaproponowano sposób integracji danych z czujnika GPS i systemu wizyjnego.
% Brak informacji o~użytej platformie obliczeniowej, rozdzielczości kamery oraz możliwości realizacji obliczeń w czasie rzeczywistym.

Summarizing the review, it is worth noting that in most works vision systems were able to process just a~few frames per second.
The authors of \cite{Lee_2018} pointed out that that this could be the reason the drone was oscillating during landing.
Moreover, in \cite{Jin_2016} one of the mentioned challenges was the use of an~energy-efficient platform to perform calculations in real-time.
In this work, we claim that the use of a~heterogeneous computing platform can address the both mentioned issues.

%Podsumowując przegląd warto zauważyć, że w większości prac system wizyjny umożliwia przetwarzanie kilku ramek na sekundę. 
%Dodatkowo autorzy pracy \cite{Lee_2018} wskazali, że może to być przyczyną występujących oscylacji.
%Ponadto w~artykule \cite{Jin_2016} jako jedno z wyzwań określono wykorzystanie efektywnej energetycznie platformy do realizacji obliczeń w czasie rzeczywistym.
%W niniejszej pracy pokazujemy, że użycie heterogenicznej platformy obliczeniowej pozwala zaadresować wskazanie problemy. 

% ------------------------------------------------------------------------------------------------------------
\section{The proposed automatic landing algorithm}
\label{sec:algorithm}

In this section we present the used hardware setup, the proposed automatic landing procedure and the landing pad detection algorithm.

% ------------------------------------------------------------------------------------------------------------
\subsection{Hardware setup}

%dron z Pixhawkiem
%Arty Z7 (FPGA + procesor)
%lidar
%radiokomunikacja bezprzewodowa do uartów
%kamera

In this research we used a~custom multirotor (hexacopter) built from the following components:
\begin{itemize}
	\item frame: DJI F550,
	\item propeller: reinforced, with the designation 9050, i.e. with a~propeller diameter equal to 9.0" (22.86~cm) and 5.0" pitch (12.7~cm),
	\item engines: DJI 2312 / 960KV controlled by 420 LITE controllers,
	\item power supply: four-cell Li-Po battery with a~nominal voltage of 14.8V (maximum 16.8V) and with a~capacity of 6750~mAh,
	\item radio equipment: FrSky Taranis X9D Plus,
	\item receiver: FrSky X8D,
	\item autopilot: 3DR Pixhawk.
\end{itemize}

% W~badaniach wykorzystano multirotor (heksakopter) zbudowany z następujących komponentów:
% \begin{itemize}
%	\item rama: DJI F550,
%	\item śmigła: wzmocnione o oznaczeniu 9050, czyli o średnicy śmigła równej
%	      9.0” (22.86cm) oraz skoku śmigła 5.0 (12.7cm), 
%	\item silniki: DJI 2312/960KV sterowane kontrolerami 420 LITE,
%	\item zasilanie: czterokomorowa bateria LiPo o nominalnym napięciu 14.8V (maksymalnym
%	      16.8V) oraz o pojemności 6750mAh,
%	\item aparatura radiowa: FrSky Taranis X9D Plus,
%	\item odbiornik: FrSky X8D,
%	\item autopilot: 3DR Pixhawk.
%\end{itemize}

\begin{figure}[!t]
	\centering
	%\hspace*{-2.5em}
	\includegraphics[width=0.8\linewidth]{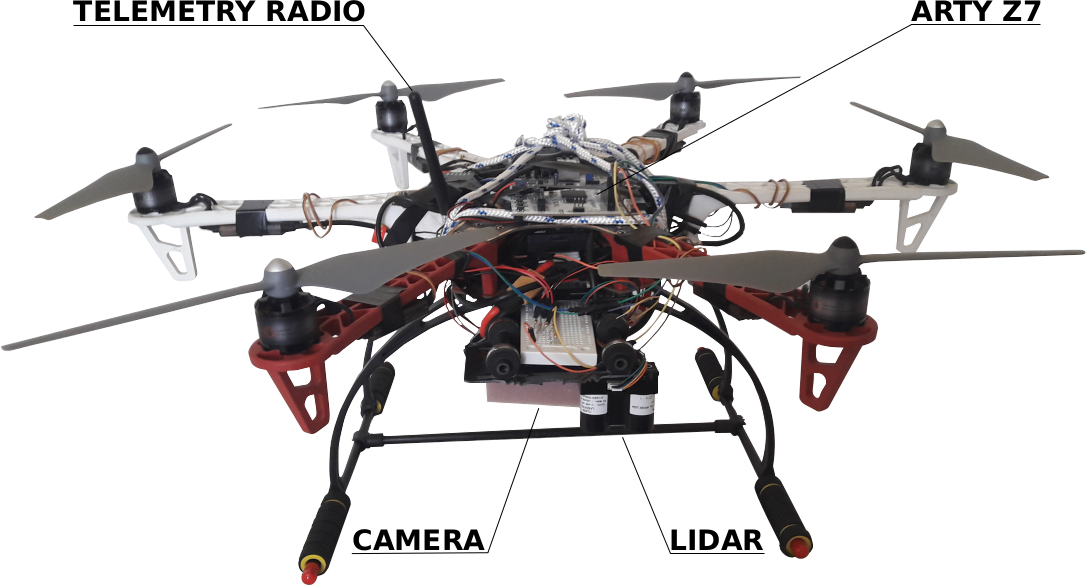}
	%\vspace*{-4em}
	\caption{The used hexacopter}
	\label{fig:uav_photo}
\end{figure}

\begin{figure}[!t]
	\centering
	% \hspace*{-2.5em}
	\includegraphics[width=0.9\linewidth]{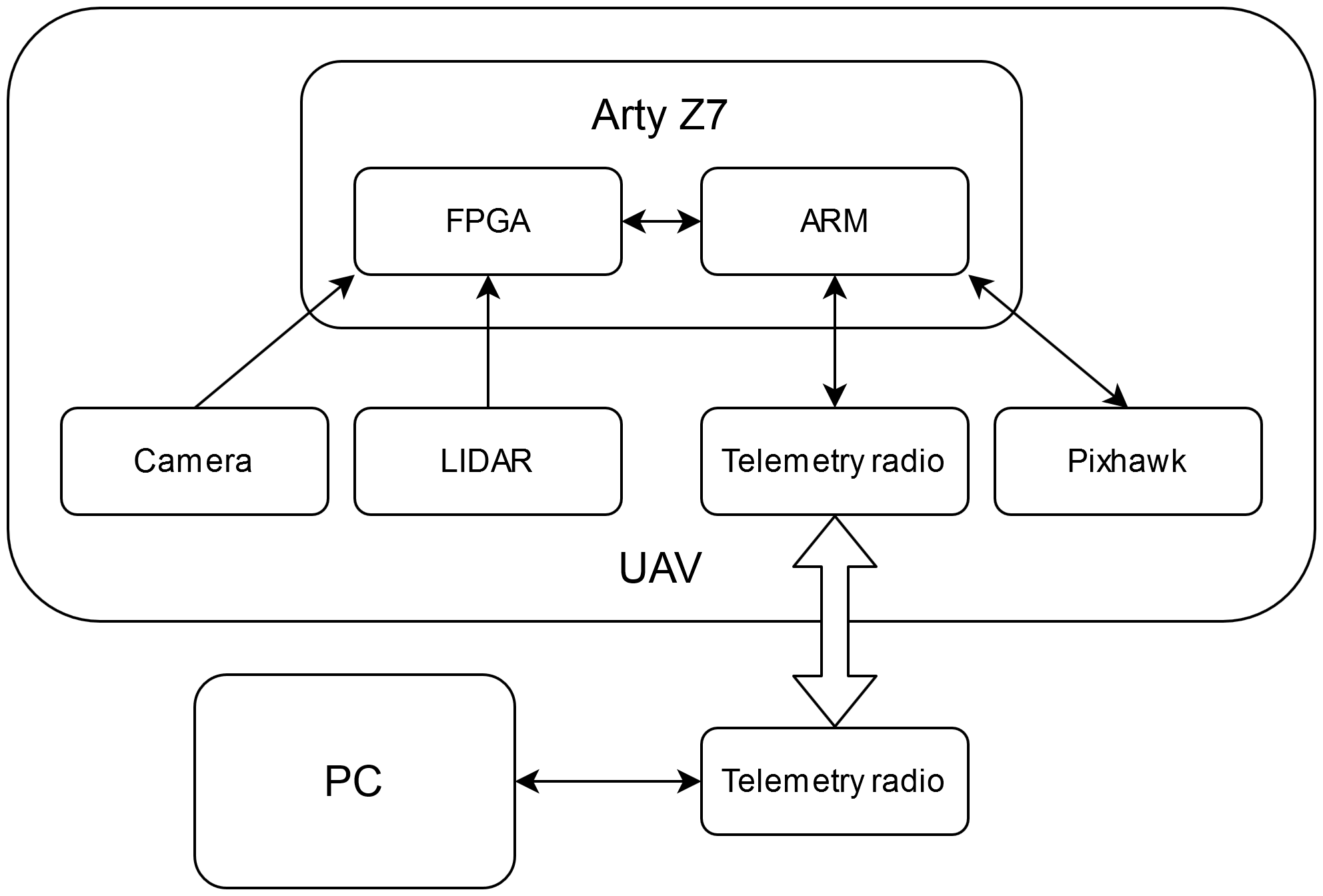}
	%\vspace*{-2.5em}
	\caption{A~simplified scheme of the proposed system}
	\label{fig:hw_diagram}
\end{figure}

The drone was adapted to the considered project.
We added a~heterogeneous computing platform Arty Z7 with Zynq-7000 SoC device.
We connected an Yi Action camera to it (as the source of a~$ 1280 \times 720 $ @ 60 fps video stream), a~LiDAR for measuring altitude above the ground level (LIDAR-Lite v3 device, SEN-14032) and a~radio transceiver module for wireless communication with the ground station (3DR Mini V2 telemetry modules).
Moreover, the Arty Z7 board was connected to the Pixhawk autopilot via UART (Universal Asynchronous Receiver-Transceiver).
The drone is shown in Figure \ref{fig:uav_photo}, while the simplified functional diagram of the proposed system is presented in Figure \ref{fig:hw_diagram}. 

%Dron został dostosowany do potrzeb projektu.
%Umieszczono na nim heterogeniczną platformę obliczeniową Arty Z7 z układem SoC Zynq-7000.
%Do niej podłączona została kamera Yi Action, stanowiąca źródło sygnału wizyjnego o~paramentach $1280 \times 720$ @ 60 fps, oraz LIDAR do pomiaru odległości od powierzchni ziemi (urządzenie LIDAR-Lite v3, SEN-14032).
%Na dronie zamontowany został także radiowy moduł nadawczo-odbiorczy do bezprzewodowej komunikacji z kontrolerem (moduły telemetrii 3DR Mini V2). %TODO dopisać. - dopisane
%Uproszczony schemat wykorzystanego systemu zaprezentowany został na rys. \ref{fig:hw_diagram}.
%Z kolei na rys. \ref{fig:uav_photo} przedstawiono użytego w projekcie drona wraz z zamontowanymi na nim urządzeniami.

\subsection{The proposed automatic landing procedure}

In the initial situation, the drone flies at an altitude of about 1.5--2~m above the ground level and the landing pad is in the camera's field of view. 
Until then, the vehicle is piloted manually or in another automated mode (using GPS-based navigation).
After fulfilling the conditions mentioned above, the system is switched into autonomous landing mode (currently implemented by a~switch on the remote controller).

%W~sytuacji wyjściowej dron leci na wysokości ok. 1.5-2m nad ziemią, a~lądowisko znajduje się w~polu widzenia kamery (założono, że kąt nie jest większy niż $45^{\circ}$).
%Do tego momentu pojazd jest pilotowany ręcznie.
%Po spełnieniu wyżej wymienionych warunków następuje przełączenie w~tryb lądowania autonomicznego (techniczne realizowane poprzez przełącznik na aparaturze radiowej). 

In the first phase, a~rough search for the landing pad is executed.
After finding it, the drone autonomously changes its position so that the centre of the marker is around the centre of the image.
In the second phase, the altitude is decreased to approx. 1~m and the drone's orientation relative to the landing pad is additionally determined.
Based on this information, the vehicle is positioned accordingly.

%W~pierwszej fazie następuje ,,zgrubne'' poszukiwanie znacznika lądowiska. 
%Po jego znalezieniu podawana jest do drona komenda zmiany pozycji w taki sposób, aby środek znacznika znajdował się w okolicach środka obrazu.
%W~drugiej fazie następuje obniżenie wysokości do ok. 1 m i~dodatkowo wyznaczana jest orientacja drona względem lądowiska.
%Na podstawie tej informacji następuje odpowiednie ustawienie pojazdu.

In the last phase, the landing is carried out.
The altitude above the landing pad is measured using the LiDAR sensor.
The position of the pad is constantly determined and corrections, if required, are made.
The drone is lowered and the engines are turned off after landing.

%W~ostatniej fazie realizowane jest lądowanie.
%Wysokość nad lądowiskiem mierzona jest za pomocą LiDAR-u.
%Pozycja lądowiska jest cały czas określana i wprowadzane są ew. korekty położenia.
%Następuje obniżenie wysokości drona i wyłączenie silników po wykonaniu lądowania.

\subsection{Landing spot detection algorithm}
\label{ssec:landing_spot_detection_algortihm}

%rgb2gray
%adaptacyjna binaryzacja
%erozja
%mediana
%dylatacja

%indeksacja
%analiza obiektów
%zestaw warunków na kółka, kwadrat, prostokąt
%pomocnicze tablice
%analiza wykrytych figur i kolejne warunki
%wyznaczenie środka obiektu i orientacji znacznika

%śledzenie?
%albo niewielkie zmiany między ramkami, albo jakieś Kalmany (ostrożnie!)

% Kluczowym elementem wykonania lądowania w trybie autonomicznym z~wykorzystaniem wizyjnego sprzężenia zwrotnego jest wykrycie samego lądowiska.
% Na podstawie analizy literatury oraz własnych eksperymentów zdecydowano się oznaczyć lądowisko w sposób przedstawiony na rys. \ref{fig:landing_strip}.
% Należy również zaznaczyć, że zaprojektowano go w~taki sposób, aby algorytmy segmentacji i~detekcji poszczególnych kształtów były możliwe do realizacji w~potokowym systemie wizyjnym w~układzie reprogramowalnym.

% Wybrany znacznik składa się z~dużego czarnego pierścienia, wewnątrz którego umieszczone zostały trzy inne figury geometryczne -- kwadrat, prostokąt i~mały pierścień.
% Największa z figur umożliwia zgrubne wykrycie lądowiska i wyznaczenie przybliżonego środka znacznika.
% Mały pierścień pozawala poprawić dokładność wyznaczenia pozycji w przypadku niewielkiej wysokości drona oraz zapobiega sytuacji, w~której duży pierścień nie byłby w~całości widoczny na obrazie.
% Z kolei celem umieszczenia kwadratu i prostokąta jest możliwość określenia orientacji całego znacznika.
% Figury te łatwo wykryć oraz od siebie odróżnić.
% Zastosowane rozwiązanie pozwala na lądowanie drona w~określonym miejscu i~z~zadaną orientacją.

\begin{figure}[!t]
	\centering
	%\hspace*{6em}
	\includegraphics[width=0.4\linewidth]{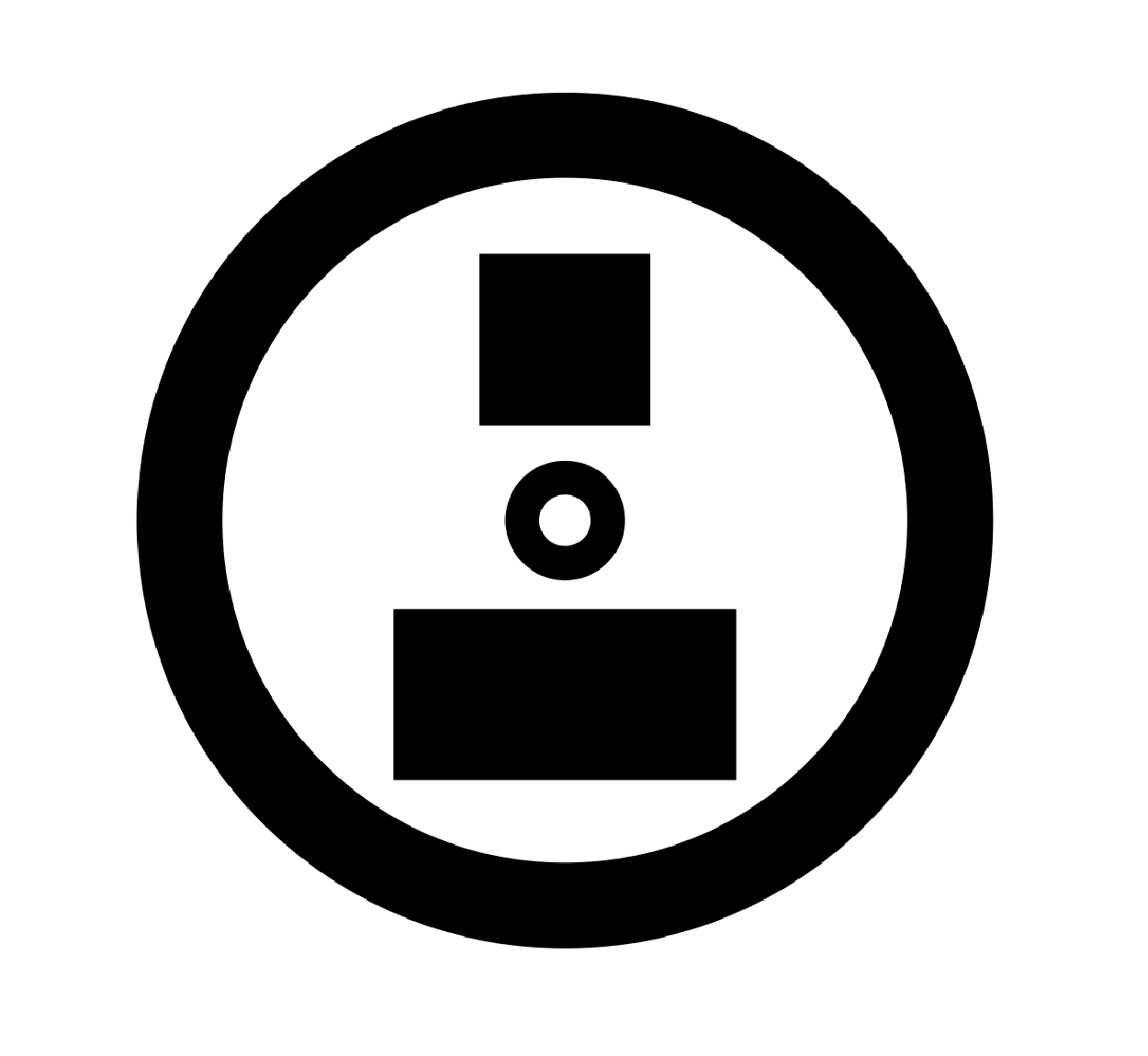}
	%\vspace*{-0em} 
	% \caption{Wybrany znacznik lądowiska}
	\caption{The used landing marker}
	\label{fig:landing_strip}
\end{figure}

The key element of the proposed autonomous landing system is the landing pad detection algorithm.
Based on the analysis of previous works and preliminary experiments, we decided to choose the landing pad marker as shown in Figure~\ref{fig:landing_strip}.
It consists of a~large black circle, inside which we placed three other geometrical figures -- a~square, a~rectangle and a~small circle.
The largest of the figures made it possible to roughly detect the landing pad and determine the approximate centre of the marker.
The small circle was used in the case of a~low altitude of the drone i.e.~when the large circle was no longer fully visible in the image.
It also allowed to improve the accuracy of the marker centre detection.
The purpose of placing the square and the rectangle was to determine the orientation of the entire marker.
Therefore, we are able to land the drone in a~certain place with a~given orientation.
Here we should note, that the used shapes are relatively easy to detect in a~vision system implemented in reconfigurable logic resources.

\begin{figure}[!t]
	\centering
	% \hspace*{-1em}
	\includegraphics[width=0.8\linewidth]{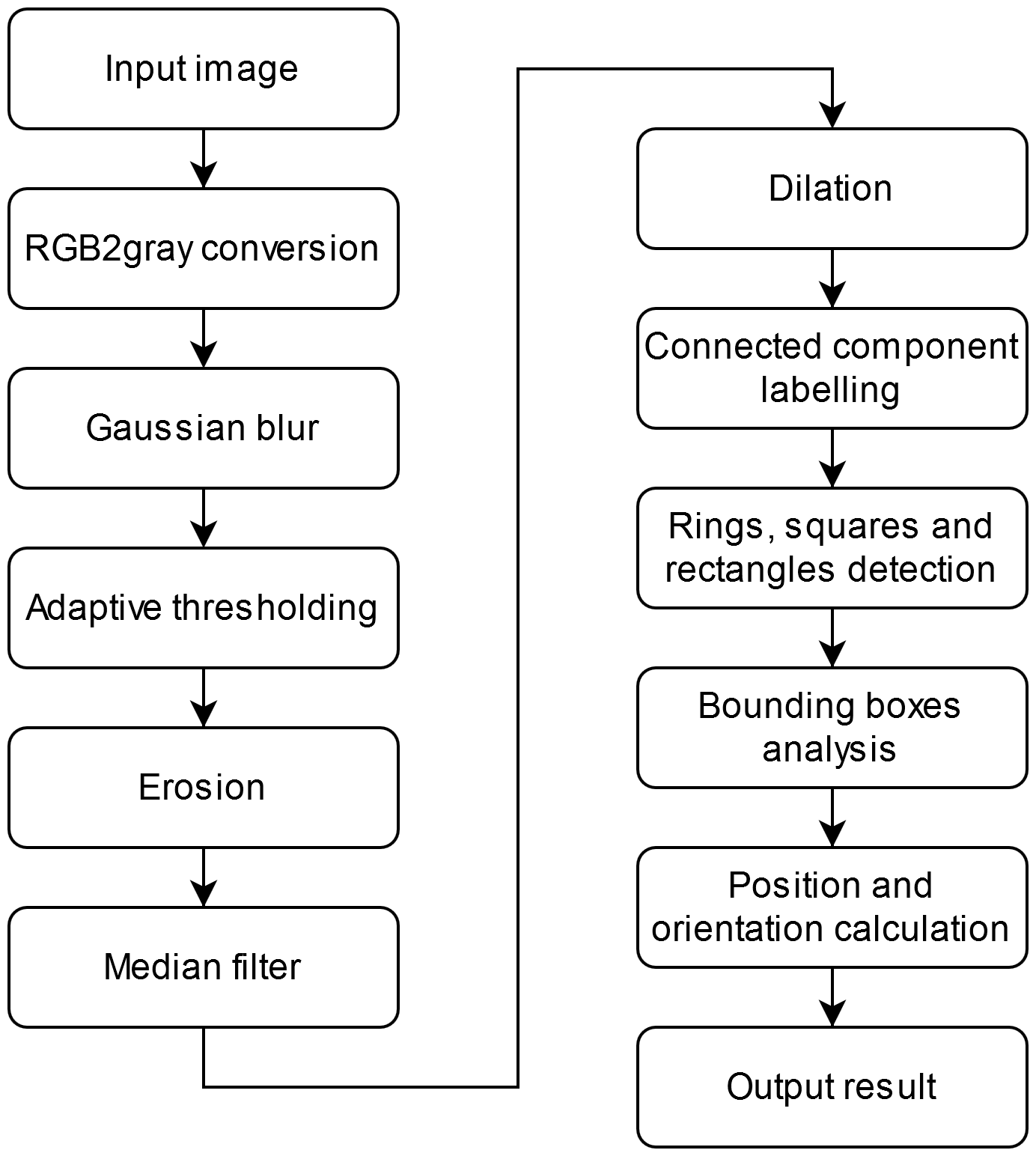}
	%\vspace*{-3em}
	% \caption{Schemat blokowy algorytmu wizyjnego}
	\caption{Simplified dataflow of the vision algorithm}
	\label{fig:block_diagram}
\end{figure}

% Uproszczony schemat blokowy całego algorytmu wizyjnego przedstawiono na rys. \ref{fig:block_diagram}.
% Do jego realizacji wykorzystano język C++ wraz z biblioteką OpenCV w wersji 4.1.0, w której dostępna jest większość podstawowych operacji przetwarzania obrazów.
% Biblioteka ta posłużyła do porównania różnych metod, doboru parametrów itp.
% Określone w ten sposób rezultaty mogły być następnie wykorzystane w czasie implementacji sprzętowej na docelowej platformie.%TODO @TK tutaj napisałem o OpenCV
% W~pierwszej kolejności rejestrowany obraz konwertowany był do odcieni szarości.
% Następnie wykonywana była adaptacyjna binaryzacja, której główną zaletą jest możliwość poprawnego działania w różnych warunkach oświetlenia, w~tym przy oświetleniu nierównomiernym.
% Obraz wejściowy dzielony był na nienachodzące okna o~rozmiarze $128 \times 128$. %TODO - zrobione
% Dla każdego z~nich wyznaczono minimalną i~maksymalną jasności piksela, a~tej podstawie określano próg binaryzacji zgodnie z~równaniem:

A~simplified block diagram of the entire vision algorithm is depicted in Figure \ref{fig:block_diagram}.
The application was prototyped in the C++ language with the OpenCV library version 4.1.0, which includes most basic image processing operations.
We provide the source code of our application \cite{SW_model_2020}. %TODO
This allowed to compare different approaches, select parameters, etc.
The obtained results could then be used during hardware implementation on the target platform.

Firstly, the recorded image is converted to greyscale.
Then a~standard Gaussian blur filtering is applied with a~$ 5 \times 5 $ kernel.
In the next step adaptive thresholding is used. 
As a~result of its use, the segmentation of the landing pad works well in different lighting conditions.
The input image is divided into non-overlapping windows of $ 128 \times 128 $ pixels.
For each of them, the minimum and maximum brightness is determined.  
Then thresholding is performed. 
The threshold is calculated according to Equation (\ref{eq:adaptivethres}).

\begin{equation} 
	\label{eq:adaptivethres}
	th = 0.25 \cdot (max - min) + min
\end{equation}
% gdzie:
where:
\begin{description}
	% \item[$th$] -- wyznaczany dla okna próg binaryzacji
	% \item[$max$] -- maksymalna jasność piksela w oknie
	% \item[$min$] -- minimalna jasność piksela w oknie
	\item[$th$] -- local threshold for the window,
	\item[$max$] -- maximum pixel brightness in the window,
	\item[$min$] -- minimum pixel brightness in the window.
\end{description}

% Drugim etapem adaptacyjnej binaryzacji jest interpolacja biliniowa progu binaryzacji.
% Dla każdego piksela określa się jego czterech sąsiadów (dla pikseli na brzegu dwóch, a~w rogu dla jednego) i~oblicza próg binaryzacji.
% Tym samym na wynikowej masce binarnej nie są widoczne granice pomiędzy analizowanymi wcześniej oknami.

\begin{figure}[!t]
	\hspace*{-1em}
	\centering
	\begin{subfigure}{0.5\linewidth}
		\centering
		%\hspace*{-2em}
		%\vspace*{1em}
		\includegraphics[width=1\textwidth]{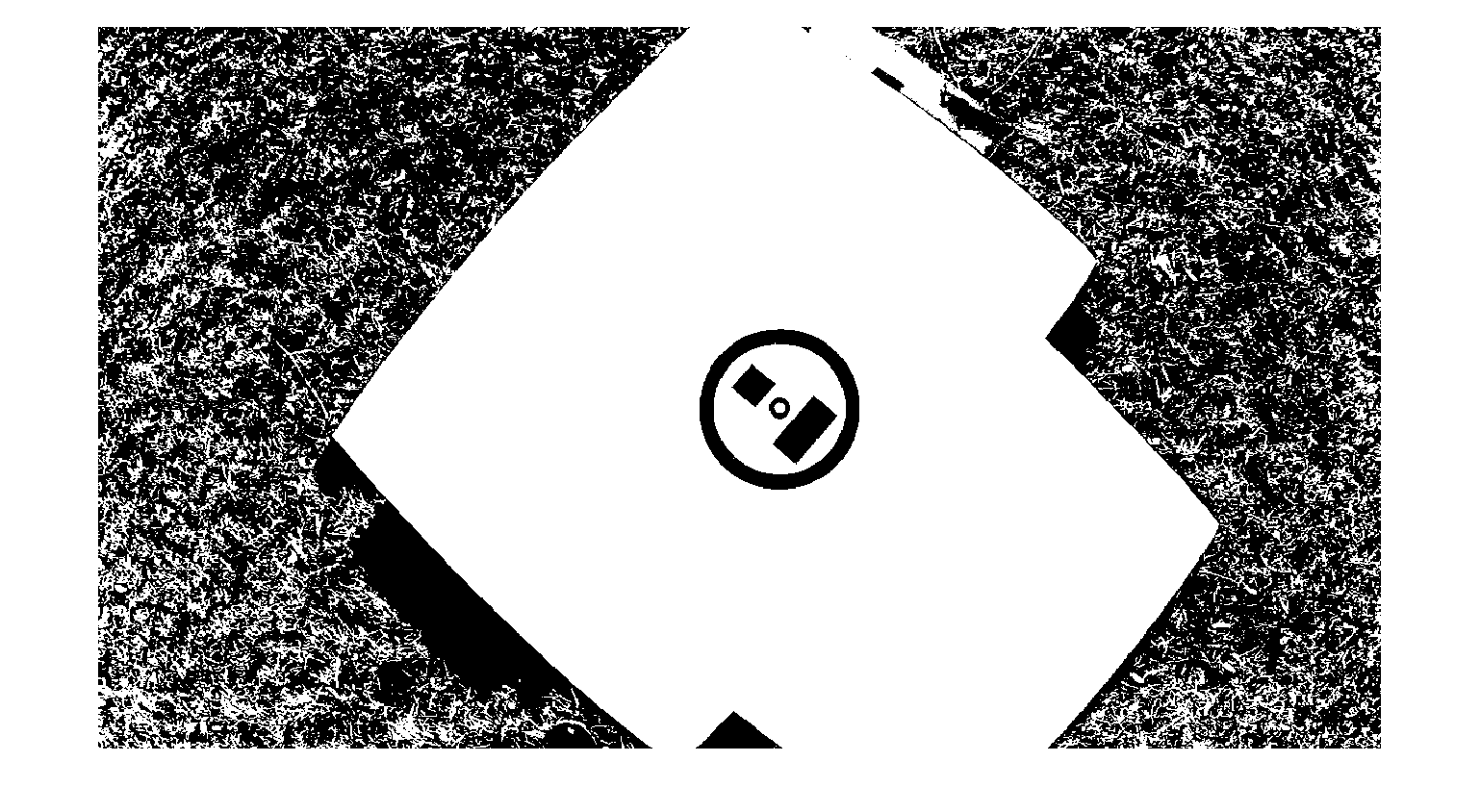}
		\vspace*{-1.5em}
		% \subcaption{Bin. globalna}\label{subfig:global_bin}
		\subcaption{Global thresholding}\label{subfig:global_bin}
	\end{subfigure}
	\begin{subfigure}{0.5\linewidth}
		\centering
		%\hspace*{-2em}
		%\vspace*{1em}
		\includegraphics[width=1\textwidth]{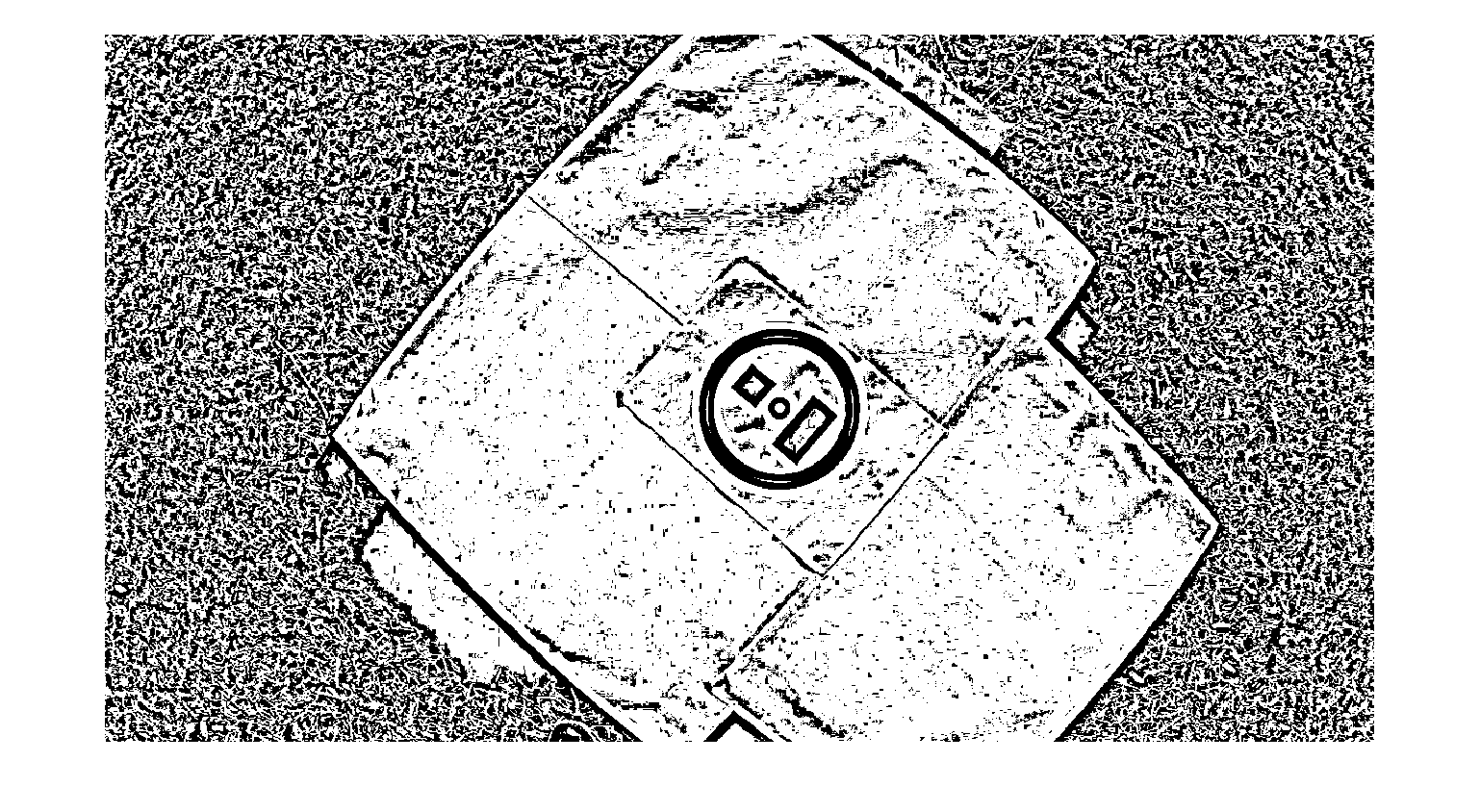}
		\vspace*{-1.5em}
		% \subcaption{Bin. lokalna}\label{subfig:local_bin}
		\subcaption{Local thresholding}\label{subfig:local_bin}
	\end{subfigure}
	\\
	\hspace*{-1em}
	\vspace*{1em}
	\centering
	\begin{subfigure}{0.5\linewidth}
		\centering
		%\hspace*{-2em}
		\vspace*{1em} \includegraphics[width=1\textwidth]{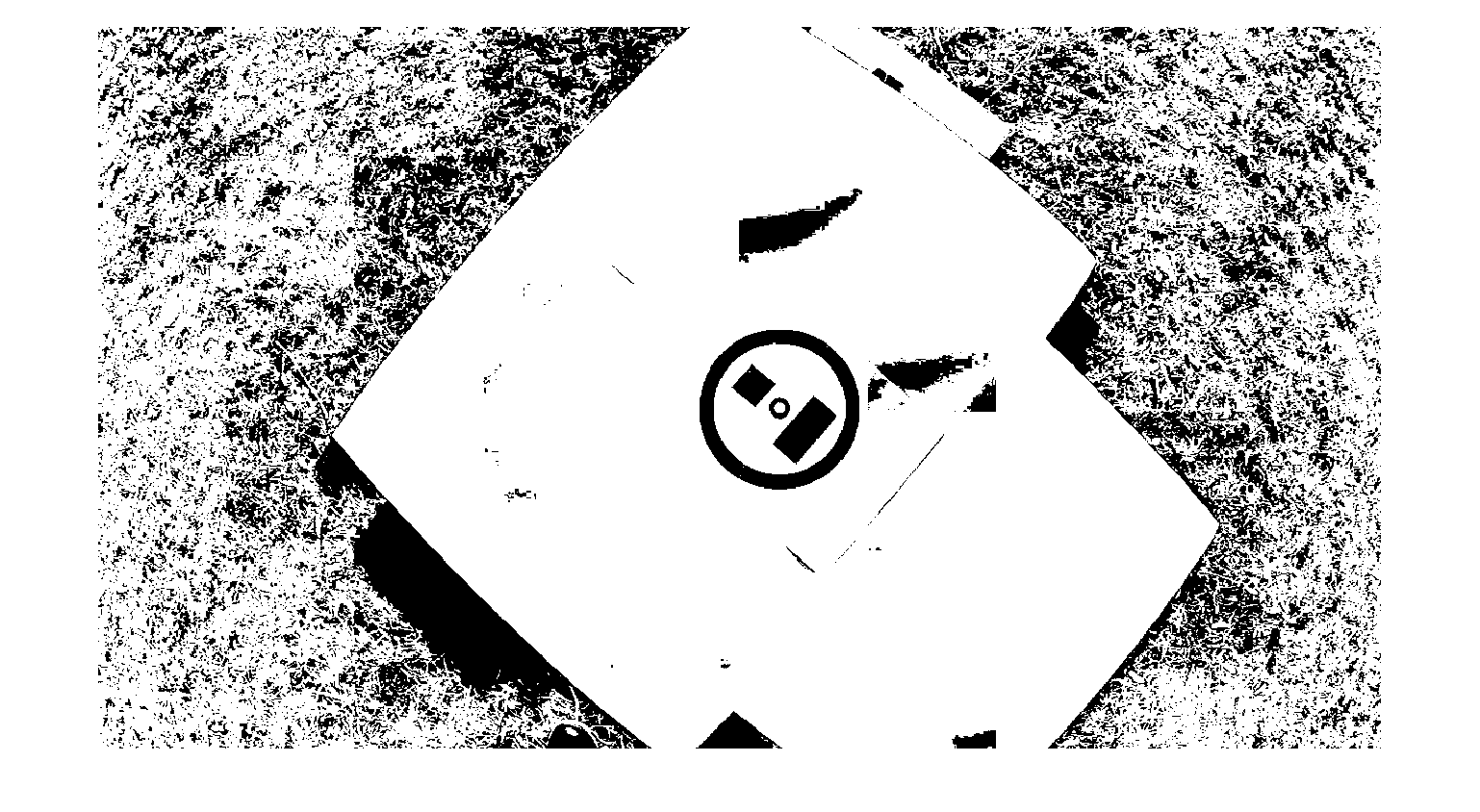}
		\vspace*{-1.5em}
		% \subcaption{Bin. w oknach}\label{subfig:adapt_bin}
		\subcaption{Adaptive in windows}\label{subfig:adapt_bin}
	\end{subfigure}
	\begin{subfigure}{0.5\linewidth}
		\centering
		%\hspace*{-2em}
		\vspace*{1em}
		\includegraphics[width=1\textwidth]{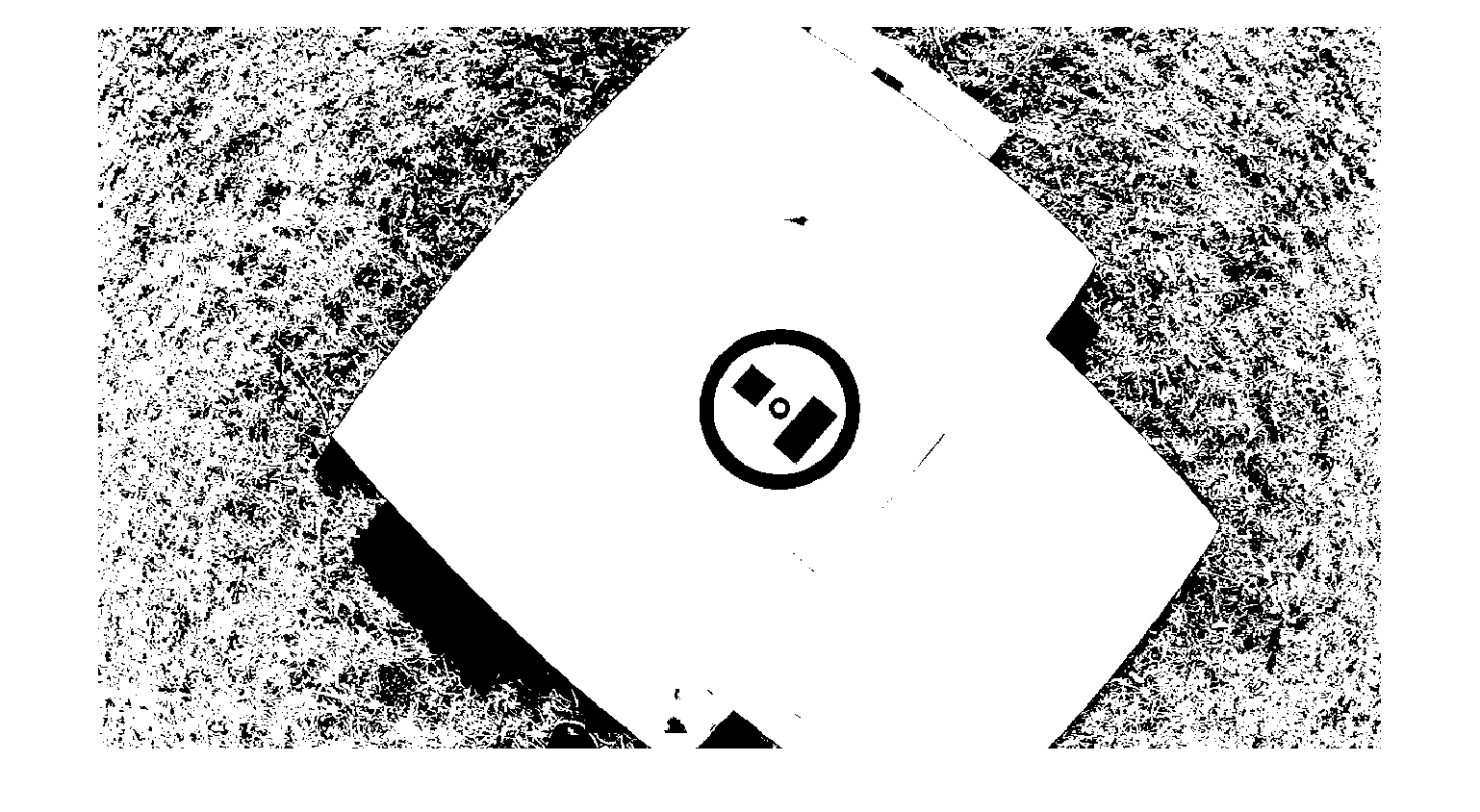}
		\vspace*{-1.5em}
		% \subcaption{Bin. z interpolacją}\label{subfig:adapt_bin_interp}
		\subcaption{Adaptive with interpolation}\label{subfig:adapt_bin_interp}
	\end{subfigure}
	% \caption{Porównanie metod binaryzacji}\label{fig:thresholding}
	\caption{Comparison of thresholding methods}\label{fig:thresholding}
\end{figure}

The second stage of adaptive thresholding is the bilinear interpolation of the threshold's value. % -- Figure \ref{subfig:adapt_bin_interp}.
For each pixel, its four neighbours are determined (for pixels on the edges two neighbours, while in the corner just one) and the binarization threshold is calculated.
As a~result, the output binary image is smoother and the boundaries between the windows are not visible.
This is presented in Figure \ref{subfig:adapt_bin_interp}.

% Warto zaznaczyć, że rozważano również alternatywne podejścia do binaryzacji tj. na podstawie progu globalnego oraz ,,w~pełni lokalną'' (próg dla każdego piksela obliczany na podstawie lokalnego otoczenia).
% Na rysunku \ref{fig:thresholding} zaprezentowano porównanie podejść dla przykładowego obrazka.
% Na podstawie analizy wyników można dojść do wniosku, że metoda adaptacyjna z interpolacją jest w tym przypadku najlepszym wyborem.
% Zwraca ona bardzo dobre wyniki i działa poprawnie w przypadku nierównomiernego oświetlenia, co daje nadzieję na największą skuteczność tej metody w czasie lotu drona.
%TODO Rysunek, komentarz. - komentarz i rysunek dodane

It is worth noting that alternative binarization approaches were also considered, i.e.~based on a~global threshold -- Figure~\ref{subfig:global_bin}, ``fully local'' (the threshold for each pixel calculated on the basis of its local neighbourhood) -- Figure~\ref{subfig:local_bin}, and adaptive in non-overlapping windows but without interpolation Figure~\ref{subfig:adapt_bin}.
Figure \ref{fig:thresholding} presents a~comparison of the described above approaches performed on a~sample image.
It can be concluded that the adaptive interpolation method is the best choice for the considered system.
It provides very good results and works correctly in case of uneven illumination.

% Kolejnym krokiem było wykonanie kilku prostych operacji kontekstowych, które umożliwiły usunięcie zakłóceń oraz drobnych obiektów.
% Najpierw obraz poddany został erozji z kontekstem 3x3.
% Następnie wykonano filtrację medianową dla rozmiaru okna 5x5, co pozwoliło na usunięcie szumu w postaci niewielkich grup pikseli.
% Na końcu wykonano dylatację dla kontekstu 3x3, aby pozostałe obiekty zachowały swoje pierwotne rozmiary.
% Przykładowy rezultat, otrzymany po wykonaniu wspomnianych filtracji, przedstawiono na rys. \ref{fig:filters}.

\begin{figure}
	\centering
	\begin{minipage}[t]{0.49\linewidth}
		\centering
		\hspace*{-1em}
		\includegraphics[width=1.1\linewidth]{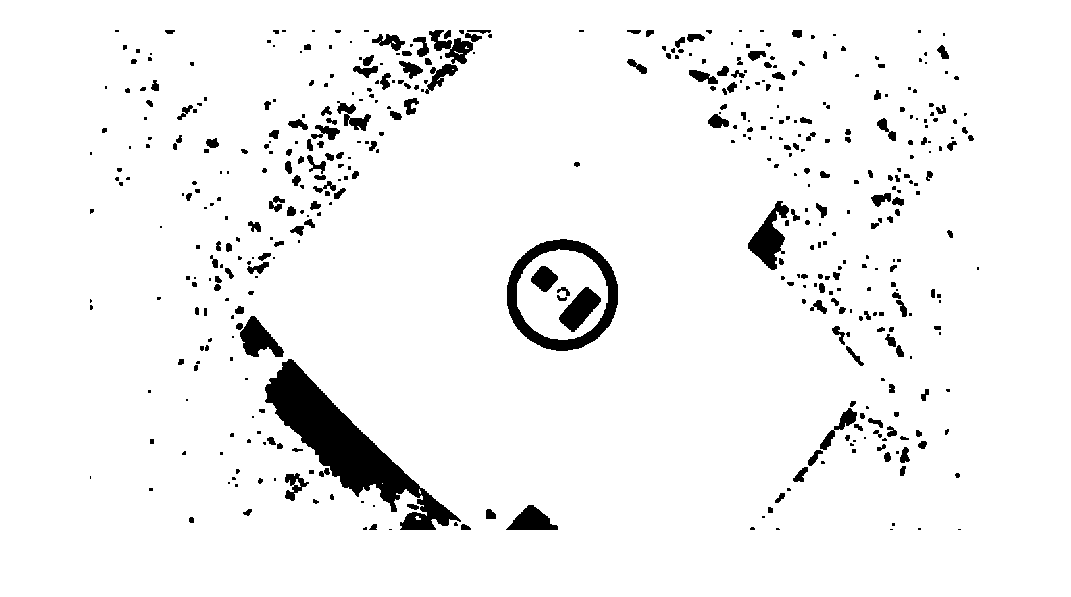}
		\captionof{figure}{Sample image after filtration}
		\label{fig:filters}
	\end{minipage}
	\hfill
	\begin{minipage}[t]{0.49\linewidth}
		\centering
		\hspace*{-1em}
		\includegraphics[width=1.1\linewidth]{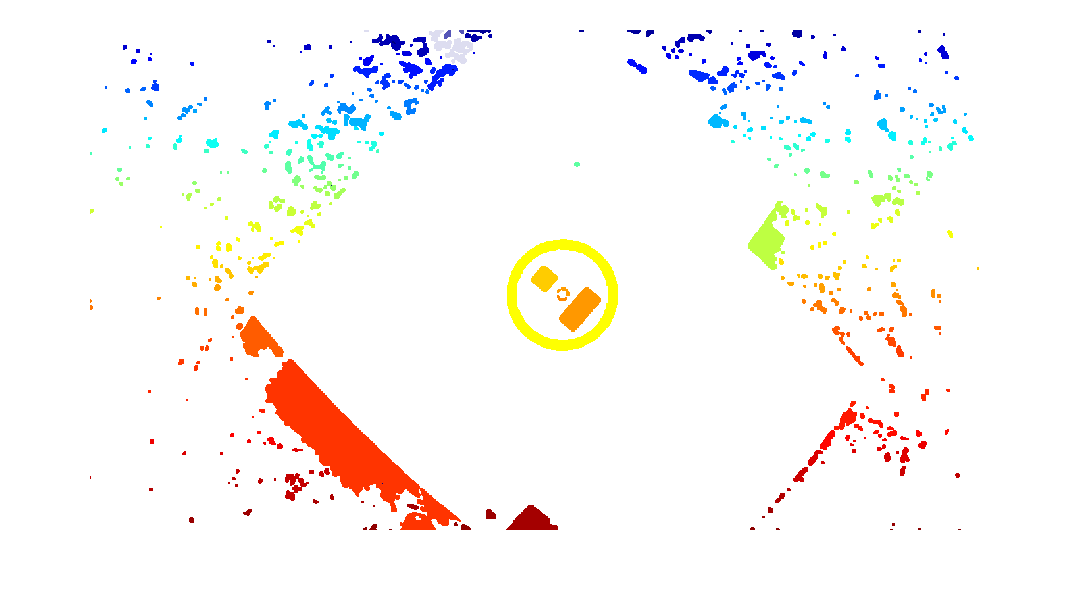}
		\captionof{figure}{Sample result of the CCL module}
		\label{fig:ccl_result}
	\end{minipage}
\end{figure}

The next steps are few simple context filtering operations used to remove small objects.
At first, erosion with a~$ 3 \times 3 $ kernel is applied.
Then median filtering with a~$ 5 \times 5 $ window is performed, which helps to remove single outliers.
Finally, dilation with a~$ 3 \times 3 $ kernel is used in order to keep the original dimensions of the remaining objects.
An~example result obtained after performing these filtrations is shown in Figure~\ref{fig:filters}.

The next step is connected component labelling.
An~example result is shown in Figure \ref{fig:ccl_result}.
As an~output: area, coordinates of the bounding box and the centroid for each object are obtained.
These values are used to calculate specific parameters, such as the size of the bounding box, its shape or the ratio of pixels belonging to the object to the total number of pixels inside it.
The analysis of these parameters allowed us to determine a~set of conditions that distinguishes the circles, squares and rectangles from other objects.
For more details please refer to the provided source code \cite{SW_model_2020}.
In addition, the expected sizes of each of the mentioned figures are determined using altitude data provided by the used LiDAR, as their dimensions in pixels depend on the distance from the camera.
The main motivation for this approach was to reduce the number of false detections.

% Aby dokładnie wykryć cały znacznik, a nie tylko pojedyncze figury geometryczne, dokonano szczegółowej analizy prostokątów otaczających obiekty.
% Kwadrat i prostokąt uznawane były za należące do znacznika, jeśli ich \textit{bounding boxy} znajdowały się wewnątrz prostokąta otaczającego duży pierścień.
% Takie podejście znacząco zmniejszyło liczbę odosobnionych figur geometrycznych. %TODO @KB Niejasne - nie chodziło o fałszywe detekcje ? - przeformułowałem; dokładnie chodziło o kwadraty i prostokąty, które należały do znacznika (bo osobny kwadrat może i nawet powinien być wykryty, ale nie powinien być uznany jako element znacznika, jeśli nie jest wewnątrz dużego pierścienia)

To accurately detect the entire marker, not just individual shapes, we performed a~detailed analysis of the bounding boxes.
The square, the rectangle and the small circle are considered as correctly detected if their bounding boxes are inside the bounding box of the large circle.
This approach has significantly reduced the number of false detections.

% Ostatecznie wyznaczano położenie i orientację wykrytego w ten sposób znacznika.
% Jego położenie na obrazie określone było poprzez centroid największego wykrytego pierścienia.
% Natomiast do wyliczenia orientacji posłużyły centroidy kwadratu oraz prostokąta.
% Najpierw wyznaczono odległość pomiędzy tymi punktami i porównano je z wielkością dużego pierścienia, aby wyeliminować niepoprawne detekcje.
% Jeżeli była ona w odpowiednim przedziale, to wyznaczany był stosunek różnicy między nimi wzdłuż osi pionowej do analogicznej różnicy wzdłuż osi poziomej.
% Wyliczone w ten sposób wartości umożliwiły określenie kąta przy pomocy funkcji $\arctan$.
% Przykładowy rezultat działania algorytmu wizyjnego przedstawiony został na rys. \ref{fig:detection_result}.
% Wykryto na nim duży pierścień, kwadrat i prostokąt, w wyniku czego dla takiej pozycji i orientacji kamery otrzymano:

\begin{figure}[!t]
	\centering
	\hspace*{-1em}
	\includegraphics[width=0.9\linewidth]{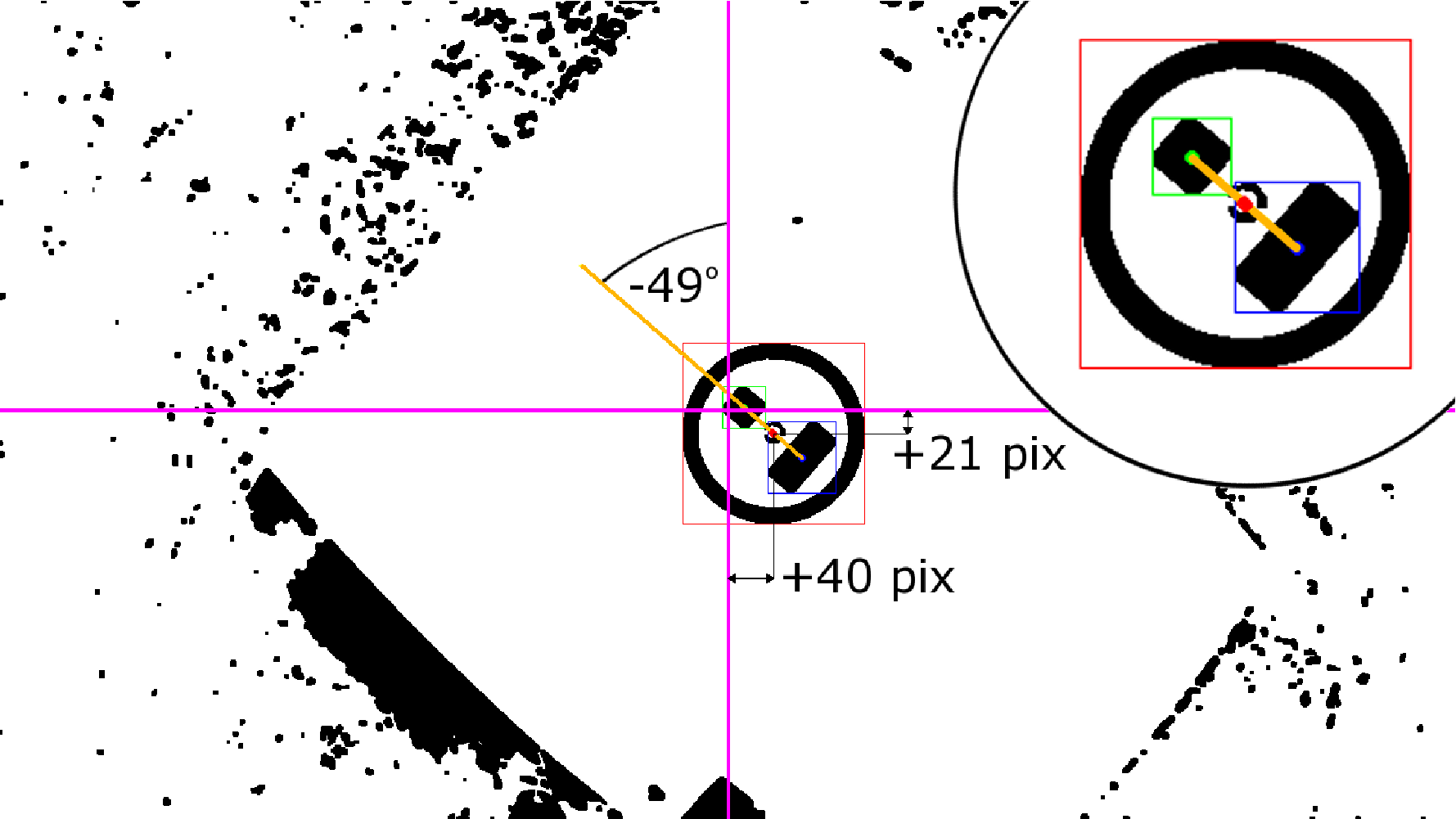}
	% \vspace*{-1em}
	% \caption{Przykładowy wynik algorytmu wizyjnego. Na czerwono zaznaczony został prostokąt otaczający pierścienia, na zielono kwadratu, a na niebiesko prostokąta. Czerwoną kropką oznaczono środek znacznika, a żółta linia wyznaczała jego orientację.}
	\caption{Sample result of the proposed vision algorithm. The centre of the camera field of view is indicated by two crossing magenta lines. The bounding box surrounding the circle is marked in red, the square in green and the rectangle in blue. The red dot marks the centre of the marker and the orange line marks its orientation. The small circle is not marked by its bounding box, because it is too small to be detected correctly. In the top right corner of the image the marker is enlarged for better visualization. The obtained drone position and orientation are also presented.}
	\label{fig:detection_result}
\end{figure}

Finally, the location and orientation of the detected marker is determined (cf. Figure~\ref{fig:detection_result}).
%Its location on the image is specified by the centroid of the largest circle detected.
The centroids of the square and rectangle are used to calculate the orientation.
Firstly, the distance between these points is determined and compared to the diametre of the large circle to avoid incorrect detections.
If that distance is in the appropriate range, the ratio of the difference between them along the Y axis to the corresponding difference along the X axis is specified.
Values calculated in this way enable to determine the angle using the $\arctan$ function.

Then the centroids of the square and the rectangle are used to calculate the position of the marker on the image.
The average values of the two mentioned centroids proved to be more reliable than the centroid of the big circle, especially in case of incomplete marker or low altitude.
However, the prior detection of the big circle is crucial as the squares and rectangles were analysed only inside it.
That means all three figures are necessary to estimate the position and orientation of the marker.
At low altitude, when the big circle is not entirely visible, the centroid of the small circle is used as the location of the landing pad centre.
The analysis of the image shown in Figure~\ref{fig:detection_result} was used in the example below to calculate the position and orientation of the drone relative to the landing pad.
The following results were obtained:
\begin{itemize}
	% \item odległość od środka obrazu w poziomie:~$+40$ pikseli
	% \item odległość od środka obrazu w pionie:~$+21$ pikseli
	% \item odchylenie względem ustalonej orientacji znacznika:~$-49^{\circ}$
	\item horizontal distance from the centre of the image:~$+40$ pixels
	\item vertical distance from the centre of the image:~$+21$ pixels
	\item deviation from the fixed marker orientation:~$-49^{\circ}$
\end{itemize}
% Po uwzględnieniu aktualnej wysokości drona i przeliczeniu wyliczonych wartości na centymetry, uzyskano:
After taking into account the current drone altitude and known dimensions of the marker, the calculated values were converted into centimetres.
The following results are obtained:
\begin{itemize}
	% \item przesunięcie w poziomie względem środka obrazu:~$6.2$ cm
	% \item przesunięcie w pionie względem środka obrazu:~$3.3$ cm
	\item horizontal offset from the centre of the image:~$6.2$ cm
	\item vertical offset from the centre of the image:~$3.3$ cm
	      
\end{itemize}

%TODO Ten rysunek większy !!! - powiększony
%TODO Marker środka. - środek jest zaznaczony kropką (widać na powiększonym znaczniku)

%TODO @TK - fragment poniżej zakomentowałem, bo to "pseudośledzenie" nie działa tak jak powinno, trzeba by trochę dłużej nad tym posiedzieć, żeby to było bardziej uniwersalne
%Aby uwzględnić zależności pomiędzy wynikami uzyskanymi na kolejnych ramkach, wprowadzono dodatkowe ograniczenia.
%Przyjęte zostało założenie, że pomiędzy dwiema następującymi po sobie ramkami zmiany położenia i orientacji znacznika są niewielkie.
%Wynika to w dużej mierze z częstotliwości próbkowania zastosowanej kamery, równej 60 klatek na sekundę.
%Takie zabezpieczenie dodatkowo zmniejsza ryzyko występowania fałszywych detekcji na pojedynczych ramkach.
%W przypadku niespełnienia tego założenia lub z powodu braku detekcji wymaganych figur geometrycznych, przyjmowana była poprzednio wyznaczona pozycja i orientacja znacznika.

%TODO Techniczne. Algorytm został napisany w z biblioteką w wersji... - napisałem to wyżej

%To take into account the relationship between the results obtained in the following frames, additional restrictions were introduced.
%We assumed that the changes in position and orientation of the marker are small between two consecutive frames.
%This is largely due to the high sampling frequency of the camera used, which was equal to 60 frames per second.
%This condition further reduces the risk of false detections on individual frames.
%In case of not fulfilling this requirement or due to the lack of detection of the required figures, the previously determined position and orientation of the marker was used.

\section{HW/SW implementation}
\label{sec:hw_sw}

%hw: odbieranie obrazu, rgb2gray, binaryzacja, erozja, mediana, dylatacja, indeksacja, lidar, wysyłanie obrazu
%sw: filtracja obiektów, analiza, komunikacja z Pixhawkiem

We implemented all necessary components of the described algorithm in a~heterogeneous system on the Arty Z7 development board with Zynq SoC device.
All image processing and analysis stages are implemented in the programmable logic of the SoC.
It receives consecutive frames from the camera, performs the required image preprocessing operations i.e.: conversion from RGB to greyscale, Gaussian low-pass filtration, median filtering, erosion, dilation and two more complex image analysis algorithms: connected component labelling (CCL) from our previous work \cite{Ciarach_2019} and adaptive thresholding.
Finally, it sends the results to the processing system (ARM processor) via AXI interface.
Moreover, in the prototyping phase the image processing results were transmitted via HDMI and visualized on a~temporarily connected LCD screen.

The adaptive thresholding algorithm, mentioned in Section \ref{ssec:landing_spot_detection_algortihm}, consists of two stages.
The first one finds the minimum and maximum values in $128 \times 128$ windows and calculates the appropriate thresholds, according to Equation \ref{eq:adaptivethres}.
The raw video stream does not contain pixel coordinates, but only pixel data and horizontal and vertical synchronization impulses, which is why the coordinates need to be calculated in additional counters implemented in programmable logic.
In the second stage, local thresholds are obtained based on the values in adjacent windows (4, 2 or 1).
It should be noted that thresholds computed for frame $N-1$, are used on frame $N$.
This approach, which does not require image buffering, is justified by the high sampling frequency of the camera and thus small differences in brightness of pixels on subsequent frames.

In addition, we use the programmable logic part to supervise the altitude measurement using LiDAR.
We use a~simple state machine to control the LiDAR in PWM (Pulse Width Modulation) mode. 
It continuously triggers the distance measurement and reads its result.
We implemented all of these modules as separate IP cores in the Verilog hardware description language.

%TODO DONE Zdanie o komunikacji (przerwania ?)
For the processing system, i.e. the ARM Cortex-A9 processor, we developed an~application in the C~programming language.
Its main task is to fuse data from the vision system and LiDAR and send appropriate commands to the Pixhawk controller via the MAVLink 2.0 protocol using UART.
In particular, it filters the objects present in the image using data obtained from the CCL module, searches for proper geometric shapes and determines the position of the drone relative to the landing pad\footnote{This is the same code as used in the software model \cite{SW_model_2020}}.
In addition, the application controls the radio module for communication with the ground station.
Thanks to this, we can remotely supervise the algorithm.
The application also uses interrupts available in the processing system, to correctly read all input data streams.
In this manner, it can stop the algorithm execution at any time, especially when an unexpected event occurs.

\begin{figure}[!t]
	\centering
	%\hspace*{-3em}
	\includegraphics[width=0.8\linewidth]{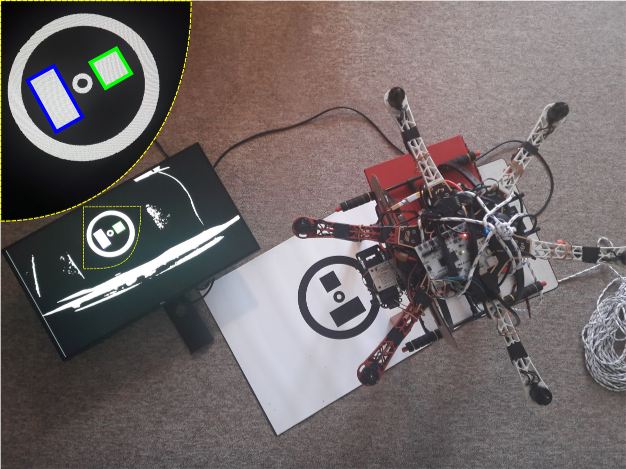}
	% \vspace*{-1em}
	% \caption{Przykładowy wynik algorytmu wizyjnego. Na czerwono zaznaczony został prostokąt otaczający pierścienia, na zielono kwadratu, a na niebiesko prostokąta. Czerwoną kropką oznaczono środek znacznika, a żółta linia wyznaczała jego orientację.}
	\caption{Working hardware-software system. The image is processed by the Arty~Z7 board on the drone at a~altitude of about 0.5~m and the detected rectangle (blue bounding box) and square (green bounding box) are displayed on the monitor.}
	\label{fig:working_system}
\end{figure}

Resource utilization of the programmable logic part of the system is presented in Table \ref{tab:utilization}.
It can be concluded that it is possible to implement improvements to the algorithm or to add further functionalities to the system.
The estimated power usage is 2.191~W. 
A~photo of the working system is presented in Figure \ref{fig:working_system}.

\begin{table}[!t]
	\centering
	\caption{PL resource utilization}
	\begin{tabular}{| c | c |}
		\hline
		{Resource} & {System}        \\	
		\hline
		LUT        & 14897 (28.00\%) \\
		\hline
		FF         & 21368 (20.08\%) \\
		\hline
		BRAM       & 24 (17.14\%)    \\
		\hline
		DSP        & 25 (11.36\%)    \\
		\hline
	\end{tabular}
	\label{tab:utilization}
\end{table}

%TODO DONE Na jakiej wysokości dron ?

\section{Evaluation}
\label{sec:evaluation}
%TODO DONE Poprawwki wynikające z ostatnich testów - poprawione

To evaluate the system, several test sequences were recorded using the Yi camera mounted on the drone.
The altitude data from the LiDAR sensor was also stored.
Diverse images of the landing pad, i.e. for drone altitude from 0~to~1.5~m, for different orientation of the marker, for its different position in the image and on different background (outside and inside) were obtained.

%Na potrzeby ewaluacji systemu nagrane zostały sekwencje testowe przy pomocy kamery zamontowanej na dronie do których dołączono dane o~wysokości pozyskane z~czujnika LiDAR.
%Sekwencje były nagrywane tak, aby uzyskać jak najbardziej różnorodne ujęcia lądowiska, tzn. dla wysokości drona od 0 do 1,5 m, dla różnej orientacji znacznika i dla różnego jego położenia na obrazie oraz na różnym tle (na zewnątrz i~wewnątrz).

At first, the marker detection rate was determined.
50 images were selected from the database, with the landing pad in different views. 
Then they were processed by the proposed algorithm.
The marker was correctly detected on 48 images, which gives a~96\% accuracy. 
The number of incorrectly detected shapes was also analysed -- not a~single shape outside the marker area has been falsely identified on the analysed images.

%W pierwszej kolejności określono skuteczność detekcji znacznika.
%Z bazy wybrano 50 obrazków, które przedstawiają lądowisko w~różnych ujęciach. %TODO czy to na pewno to 50 
%Następnie podano je na wejście algorytmu wizyjnego.
%Znacznik został poprawnie wykryty na 44 obrazach, co przekłada się na skuteczność detekcji 88\%. %TODO @KB poprawić
%Sprawdzona została także liczba niepoprawnie wykrytych figur geometrycznych -- na 50 ramkach zaznaczona została jedynie 1 figura niebędąca elementem znacznika.

Secondly, the marker centre estimation accuracy was evaluated.
The position returned by the algorithm was compared with a~reference one (selected manually).
The differences for the horizontal and vertical axes were calculated.
Then the obtained results were averaged separately for each axis. 
For the horizontal axis the deviation was 0.19 pixels, while for the vertical axis 0.67 pixels.

%Drugim etapem ewaluacji było porównanie wyznaczonej pozycji środka znacznika, czyli docelowego miejsca lądowania, z pozycją referencyjną, określoną manualnie.
%Dla tych obrazów, dla których wykryto lądowisko, określono różnice względem wartości referencyjnej w poziomie i w pionie.
%Następnie otrzymane wyniki uśredniono dla każdej z osi, dzięki czemu uzyskano średni błąd określenia pozycji znacznika.
%Dla osi poziomej odchylenie wynosiło 0.52 piksela, natomiast dla osi pionowej 0.91 piksela. 

\begin{figure}[!t]
	\centering
	\hspace*{-1em}
	\includegraphics[width=0.9\linewidth]{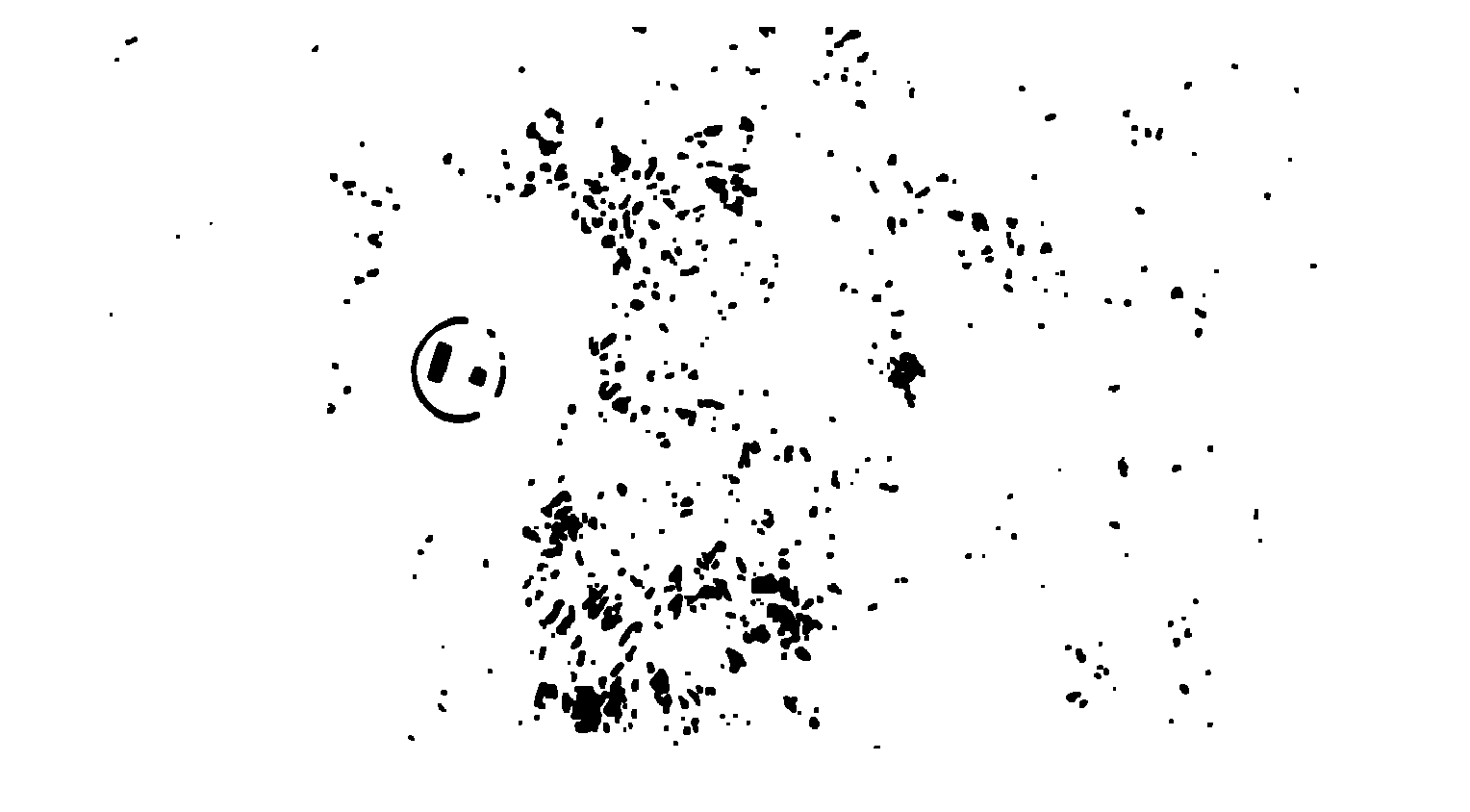}
	% \vspace*{-1em}
	% \caption{Przykładowy wynik algorytmu wizyjnego. Na czerwono zaznaczony został prostokąt otaczający pierścienia, na zielono kwadratu, a na niebiesko prostokąta. Czerwoną kropką oznaczono środek znacznika, a żółta linia wyznaczała jego orientację.}
	\caption{Sample frame without correct marker detection. In this example the circle after filtration is separated into several pieces, so the defined detection conditions are not fulfilled.}
	\label{fig:bad_detection}
\end{figure}

The analysis of the presented results allows to draw the following conclusions about the performance of the vision algorithm.
The percentage of frames with a~correctly detected marker is high, but some cases turned out to be problematic.
These were mainly images, in which the marker was far from the centre of the camera's field of view or separated into several fragments. 
%This caused its distortion and thus the failure to meet the conditions for the detection of particular figures -- this situation is presented in Figure \ref{fig:bad_detection}.
The last mentioned case caused the failure to meet the conditions for the detection of particular shapes -- this situation is presented in Figure \ref{fig:bad_detection}.
An~attempt was made to solve that problem by using a~less restrictive set of conditions.
However, this in turn resulted in false detections, which was undesirable.
Analysing the marker centre estimation result it can be concluded that this value has been determined with high accuracy (below 1 pixel), which enables precise drone navigation.

%Na podstawie przedstawionych wyników można wyciągnąć następujące wnioski na temat działania algorytmu wizyjnego.
%Procentowy udział ramek z poprawnie wykrytym znacznikiem jest wysoki, jednakże niektóre przypadki okazały się zbyt dużym wyzwaniem.
%Były to głównie obrazy, na których znacznik znajdował się daleko od środka, co powodowało jego zniekształcenie i tym samym niespełnienie warunków na detekcję poszczególnych figur.
%Podjęto próbę rozwiązania powyższego problemu poprzez zastosowanie mniej restrykcyjnego zestawu warunków do detekcji figur.
%Jednak w takim przypadku stwierdzono wzrost liczby fałszywych detekcji.

%Analizując rezultaty określania pozycji lądowisko w porównaniu z wartościami referencyjnymi, można dojść do wniosku, że pozycje lądowiska wyznaczone zostały z dużą dokładnością (poniżej 1 piksela), dzięki czemu możliwa jest precyzyjna nawigacja drona.

%TODO Inne zdjecie - dodane, jest ok?

% ------------------------------------------------------------------------------------------------------------
% TYLKO DO REPO
\vspace{1cm}
\section{Conclusion}

In this paper a~hardware-software control system for autonomous landing of a~drone on a~static landing pad was presented.
The use of a~Zynq SoC device allowed to obtain real-time processing of a~$1280 \times 720$ @ 60 fps video stream. 
The estimated power utilization of the system is 2.191~W. 
The designed algorithm has an~accuracy of 96\%. 
Unfortunately, due to the prevailing epidemic and the associated restrictions, it was not possible to test the fully integrated system (i.e. to perform a~fully autonomous landing based on the proposed vision algorithm).
This will be the first step of further work on the system and an~opportunity to gather a~larger database of test sequences in various conditions.

%W artykule zaprezentowano sprzętowo-programowy system do autonomicznego lądowania drona na statycznym lądowisku.
%Wykorzystanie układu Zynq SoC pozwoliło uzyskać przetwarzanie strumienia o~rozdzielczości 1280x720 @ 60 fps w~czasie rzeczywistym.
%Zaprojektowany algorytm charakteryzuje się skutecznością na poziomie 94 \%. %TODO.
%Niestety z~powodu panującej epidemii i~związanych z~nią ograniczeń, nie udało się wykonać testów w~pełni zintegrowanego systemu (tj. przeprowadzić w~pełni autonomicznego lądowania bazującego na zaproponowanym algorytmie wizyjnym).
%Będzie to pierwszy element dalszych prac nad systemem oraz okazja do zgromadzenia większej bazy sekwencji testowych w~różnorodnych warunkach.

The next step will be landing on a~mobile platform -- moving slowly, quickly or imitating the conditions on water / sea (when the landing pad sways on waves).
In the vision part, it is worth considering using a~camera with an~even larger viewing angle and evaluate how it affects the algorithm.
In addition, the methods that allow to distinguish the marker from the background -- like RTV (Relative Total Variation) from work \cite {Huang_2019}, should be considered.
Another option is to use an~ArUco marker, although implementing its detection in a~pipeline vision system seems to be a~greater challenge.
Moreover, adding a~Kalman filter (KF) to the system should increase reliability if detection errors occur incidentally on some frames in the video sequence.
Additionally, the fusion of a~video and IMU (Inertial Measurement Unit) data (from the Pixhawk flight controller) should be considered. %TODO ew. coś dodać mądrego

%Następnym krokiem będzie lądowanie na ruchomej platformie -- poruszającej się wolno, szybko lub też imitującej warunki panujące na wodzie/morzu (kołysanie się).
%W części wizyjnej warto rozważyć zastosowanie kamery o~jeszcze większym kącie widzenia, przy czym należy sprawdzić jaki to będzie miało wpływ na zastosowany algorytm.
%Dodatkowo warto rozważyć metody pozwalające wyróżnić marker z tła -- jak RTV z~pracy \cite{Huang_2019}.
%Inną opcją jest zastosowanie markera typu AruCo, choć implementacja jego detekcji w potokowym systemie wizyjnym stanowi większe wyzwanie.
%Dodanie do systemu filtru Kalmana (KF) powinno zwiększyć niezawodność w przypadku problemów z detekcją znacznika na pojedycznych ramkach z sekwencji.
%Dodatkowo można rozważyć fuzję danych z IMU z kontrolera lotów Pixhawk.



\section*{Acknowledgement}
%TODO Numer badań statutowych.
%Removed for blind review.
The work was supported by AGH project number 16.16.120.773. The authors would like to thank Mr. Jakub Kłosiński, who during his bachelor thesis started the research on landing spot detection and Mr. Miłosz Mach, who was the constructor of the used drone.

% TYLKO DO REPO
\vspace{2cm}



\bibliographystyle{IEEEtran}
\bibliography{2020_mmar_bswgk.bib}

% 	%\bibitem{IEEEhowto:kopka}
% 	%H.~Kopka and P.~W. Daly, \emph{A Guide to \LaTeX}, 3rd~ed.\hskip 1em plus
% 	%  0.5em minus 0.4em\relax Harlow, England: Addison-Wesley, 1999.
% 	\bibitem{Ciarach2019}
% 	P.~Ciarach, M.~Kowalczyk, D.~Przewłocka, T.~Kryjak, Real-Time FPGA Implementation of Connected Component Labelling for a 4K Video Stream In: Applied Reconfigurable Computing, 2019. ISBN 978-3-030-17227-5.

% 	\bibitem{Xu_2018}
% 	Xu, Cheng and Tang, Yuanheng and Liang, Zuotang and Yin, Hao, UAV Autonomous landing algorithm based on machine vision, IEEE 4th Information Technology and Mechatronics Engineering Conference (ITOEC), pp. 824--829, 2018.

% \end{thebibliography}

\end{document}